\documentclass[11pt]{article}

\usepackage[final]{acl}
\usepackage{times}
\usepackage{latexsym}
\usepackage[T1]{fontenc}
\usepackage[utf8]{inputenc}
\usepackage{microtype}
\usepackage{inconsolata}
\usepackage{graphicx}

\usepackage{algorithm}
\usepackage{algpseudocode}
\usepackage{listings}
\usepackage{multirow}
\usepackage{xcolor}
\usepackage{booktabs}
\usepackage{tabularx}

\usepackage{tcolorbox}
\usepackage{array}
\newcolumntype{L}[1]{>{\raggedright\arraybackslash}p{#1}}

\newcommand{\methodlink}[2]{\hyperlink{cite.#2}{\textbf{#1}}\nocite{#2}}

\newtcolorbox{promptbox}[1][]{
    colback=gray!5,
    colframe=gray!100,
    fonttitle=\bfseries\small,
    title=#1,
    boxrule=0.5pt,
    left=4pt, right=4pt, top=4pt, bottom=4pt,
    fontupper=\small\ttfamily\raggedright
}

\title{PatientHub: A Unified Framework for Patient Simulation}

\author{Sahand Sabour\quad TszYam NG\quad Minlie Huang
\\ \\
The CoAI Group, DCST, Institute for Artificial Intelligence, Tsinghua University, Beijing, China \\ \\
\texttt{sahandfer@gmail.com}, \texttt{aihuang@tsinghua.edu.cn}}

\begin{document}
\maketitle
\begin{abstract}
As Large Language Models increasingly power role-playing applications, simulating patients has become a valuable tool for training counselors and scaling therapeutic assessment. 
However, prior work remains fragmented: existing approaches rely on incompatible, non-standardized profiles, prompts, and evaluation metrics, hindering reproducibility, fair comparison, and reuse. 
We introduce PatientHub, a unified and modular framework that standardizes the creation, simulation, and evaluation of LLM-based patients.
Our framework provides 16 patient simulators, a graph-based orchestrator for multi-turn, multi-session interactions, and a configurable LLM-as-a-judge evaluator that supports multiple rubric types.
Via our command-line interface, users can generate patient profiles, run simulations, and apply rubric-driven evaluation at the turn and session level. 
To demonstrate PatientHub's utility, we compare several supported simulators under a shared interaction protocol and showcase its extensibility by prototyping a new simulator variant with minimal additional code.
By consolidating existing work into a single reproducible pipeline, our framework eliminates much of the infrastructure overhead that currently fragments this research and accelerates the development of new methods. 
Our code and data are publicly available via \url{https://github.com/Sahandfer/PatientHub}.
\end{abstract}

\section{Introduction}
Mental health disorders such as depression and anxiety are prevalent and create a sustained need for timely, accessible, and appropriate support.
Yet, access to such support remains limited due to severe workforce constraints: the global median is only 13 mental health workers per 100{,}000 people, with rates considerably lower in low-income countries \cite{who2025atlas2024}. 
Meeting this need at scale requires progress along two complementary directions: training more counselors and peer supporters \cite{steenstra2025scaffolding, wang-etal-2024-patient, louie2025llmsimulated}, and developing LLM-based supporters that provide scalable, low-cost support under appropriate safeguards \cite{sabour2023chatbot, chen-etal-2023-soulchat, zhou-etal-2025-crisp}. 
However, both directions face practical bottlenecks.

First, many training programs still rely heavily on static materials (e.g., textbooks) that cannot provide interactive practice or feedback \cite{lin2026candormd}.
A viable alternative to this is standardized patients (i.e., trained actors), in which students practice necessary skills (e.g., active listening) in low-risk settings; this approach can be effective but is costly and difficult to scale \cite{Yao2020Toward}.
Moreover, it is highly challenging to diversify the patient population across different demographic and cultural backgrounds and symptom profiles.

Second, data scarcity and safety risks constrain the development of LLM-based supporters \cite{yang-etal-2025-cami, archiwaranguprok2025simulating}. 
Creating publicly available, diverse, high-quality therapy conversations for training LLMs is difficult because of privacy concerns, ethical constraints, and the potential cost of expert involvement.
In addition, rigorously assessing LLM-based supporters requires a safe simulated environment, as involving real human users carries safety risks (e.g., failures can lead to harmful advice), which makes large-scale evaluation challenging.

A growing line of work is leveraging LLM-based simulated patients for training, evaluation, and data generation \cite{lee-etal-2024-cactus, lee2025psyche, lee-etal-2025-adaptive}.
Compared with standardized patients, these simulators offer lower marginal cost, greater personalization, and rapid scaling across diverse cases.
In addition, they have shown strong potential for generating large-scale, high-quality synthetic datasets for training and evaluating LLM-based supporters \cite{yang-etal-2025-cami, zhao-etal-2024-esc, liu2025enhanced}.

However, existing work remains scattered across papers and codebases, with substantial variation in how simulated patients are specified (e.g., profile structures), how conversations are orchestrated (single- vs. multi-turn; single- vs. multi-session), and how responses/outputs are evaluated (e.g., automatic evaluation, expert judgments, or task-specific rubrics). 
In principle, this diversity is a strength, as it enables researchers to explore various applications of simulated patients.
Yet, this fragmentation makes it difficult to reproduce prior work, compare results across different methods, and reuse existing components (e.g., profiles, prompts, evaluation rubrics). 
These capabilities are essential for building reliable benchmarks, identifying which design choices matter, and facilitating the development of novel methods by reducing duplicated effort.

Towards this end, we present PatientHub, a unified and modular framework that standardizes the creation, simulation, and evaluation of simulated patients. 
PatientHub provides a common interface for implementing diverse mental health agents, a graph-based orchestrator that composes multi-turn interactions as configurable events, and a rubric-driven LLM-as-a-judge evaluator for downstream analysis.
Through our command-line interface, researchers can generate patient profiles, run simulations across heterogeneous methods, and apply standardized evaluation, while extending the framework with their own profiles, prompts, and metrics.
Our contributions are summarized as follows:
\begin{itemize}
    \item Unify 16 patient simulators, covering a broad range of existing approaches, under a shared abstraction of agents, events, generators, and evaluators, enabling fair cross-method and cross-model comparison that was previously hindered by incompatible codebases.
    \item Provide a configurable LLM-as-a-judge evaluator supporting four rubric paradigms at both the turn and session levels.
    \item Support rapid development of new methods, which we demonstrate by prototyping a PATIENT-$\Psi$ variant with minimal code.
    \item Release the framework (MIT license) with documentation and an installable package, lowering the barrier to entry.
\end{itemize}

\begin{figure*}
    \centering
    \includegraphics[width=\linewidth]{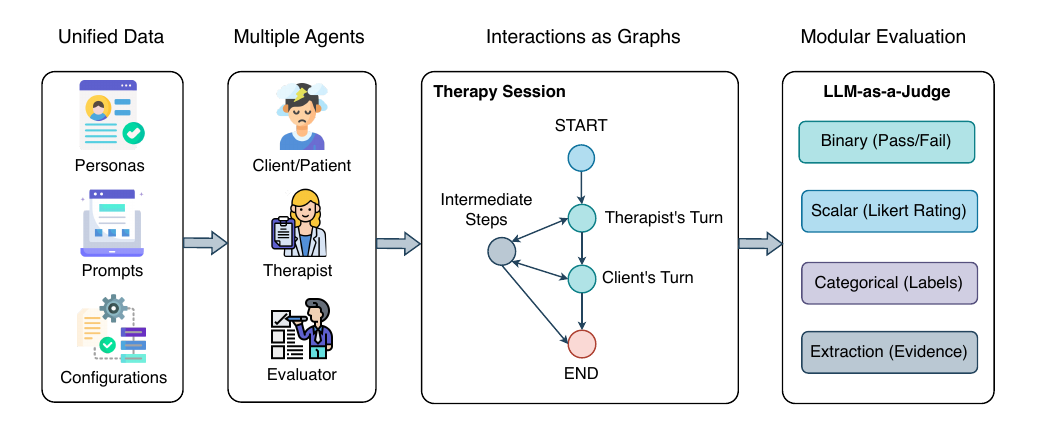}
  \caption{An Overview of PatientHub. 
  Our framework includes a unified collection of personas, prompts, and configurations from existing work; support for several agent roles for interactions; orchestration of dialogue flows/events as directed graphs; and cross-method benchmarking through a modular LLM-as-a-Judge Evaluator supporting four fundamental evaluation paradigms.
  }
    \label{fig:framework}
\end{figure*}

\section{Related Work}
\subsection{Patient Simulation Methods}
LLM-based patient simulation has emerged as a flexible alternative to traditional rule-based simulators, leveraging modern LLMs' role-playing capabilities.
Prior work varies primarily in how patient behavior is specified (e.g., profile design), what interaction protocol is assumed (e.g., single- vs. multi-session), and how desired attributes (e.g., clinical faithfulness) are enforced.
Existing methods are primarily prompt-based, conditioning generation on structured profiles and explicit constraints \cite{wang-etal-2024-patient, louie-etal-2024-roleplay, wang2024towards}, and may incorporate explicit latent states or turn-level updates to reduce drift over long conversations \cite{yang-etal-2025-consistent, wang-etal-2025-annaagent}.
Fewer approaches directly train a patient simulator; to the best of our knowledge, Eeyore \cite{liu-etal-2025-eeyore} remains the only method to date to fine-tune a depression simulator via expert-in-the-loop supervision and preference optimization.
Task settings also impact design, as simulators are often designed to meet specific clinical needs.
Simulators for screening and psychiatric assessment ground responses in symptom checklists or interview schemas \cite{wang2025talkdep, lee2025psyche, liao2024automatic}, whereas counseling-oriented simulators emphasize human-like variability \cite{du-etal-2025-llms, zhang2026human}.
Despite this diversity, these methods are built in isolation across incompatible codebases, which PatientHub consolidates under a shared abstraction.

\subsection{Applications of Patient Simulation}
Patient simulators provide repeatable, low-risk role-play settings for training and assessment.
In clinical education, they offer a scalable alternative to standardized patients, enabling trainees to practice essential skills such as empathy and active listening \cite {steenstra2025scaffolding, wang-etal-2024-patient} and providing contextualized feedback \cite{louie2025llmsimulated, campillos2021lessons, holderried2024language}.
Moreover, they serve as standardized testbeds for benchmarking supporter agents while avoiding the risks and cost of real-user evaluation \cite{sabour2025human, bi-etal-2025-magi, chiu2024computational}.
Simulated interactions also enable the creation of synthetic datasets for training and analysis \cite{yang-etal-2025-cami, lee-etal-2024-cactus, mandal2026magnet}.
Across these applications, rubric-driven \emph{LLM-as-a-judge} evaluation has become the practical mechanism for measuring efficacy, realism, and safety \cite{liao2024automatic, emerson-etal-2025-automated, yosef2024assessing, zhu2025psiarena}.
Yet, rubrics and protocols differ across studies, motivating the standardized evaluator that PatientHub provides.

\section{PatientHub}
PatientHub is a modular framework designed for the standardized creation, simulation, and evaluation of patient-centered interactions. 
Rather than specifying a single fixed design, we define a small set of composable abstractions (including agents, events, generators, and evaluators), which enable researchers to generate comparable simulations across diverse healthcare settings, from psychiatric assessment to CBT counseling.
Concretely, users can specify (i) who participates in an interaction, (ii) how the interaction unfolds, and (iii) how its inputs and outputs are evaluated, then run the entire pipeline from a single command-line interface.

\subsection{Core Components}
\paragraph{Clients.}
Client agents simulate patients in healthcare or counseling interactions, spanning both diagnosed patients and individuals seeking support. 
Each client is specified by (i) a persona profile, (ii) a response-generation specification constraining output, and optionally (iii) an internal latent state that evolves over the dialogue to model emotional shifts or symptom progression. 
PatientHub currently supports 16 simulators from existing work (Table~\ref{tab:supported_clients}); we reused the original prompts and profiles verbatim from the published papers or codebases, and only re-implemented the surrounding orchestration code to fit our abstractions.

\paragraph{Therapists.}
Therapist agents serve as counterparts to clients and use the same high-level interface. 
Alongside a prompt-driven therapist with optional chain-of-thought and a rule-based ELIZA \cite{weizenbaum1966eliza}, our framework also includes methods such as CAMI \cite{yang-etal-2025-cami} for motivational interviewing. 
Researchers can vary the therapist similar to the client, using the same profiles, events, and evaluators.

\paragraph{Events.} 
Events orchestrate agent interactions, represented as graphs that specify protocols for turn-taking, termination criteria, and session structure.
This design enables researchers to define workflows aligned with their experimental objectives.
For instance, as shown in Figure \ref{fig:framework}, the \emph{Therapy Session} event runs as a finite-state workflow, including initialization, therapist turn, client turn, and an end/reminder branch; the event is terminated upon reaching the turn limit or an explicit end signal, and all turns are logged to a unified JSON schema for downstream analysis.

\paragraph{Generators.} \label{sec:generators}
Generators synthesize the client profiles that drive simulations, constructing them from seed sources such as conversation excerpts, clinical questionnaires, or structured disorder descriptions \cite{wang2024towards, wang-etal-2025-annaagent, lai2026patientzero}.
By controlling the distribution of generated profiles, researchers can scale simulations across diverse demographics and symptom presentations reproducibly.

\paragraph{Evaluators.}\label{sec:evaluators}
PatientHub provides a standardized \emph{LLM-as-a-judge} abstraction that covers two judges: \textit{profile judge} for generated profiles and \textit{conversation judge} for full conversations, applied at the session or turn level.
Either judge can run online during a simulation or offline on saved logs.
Each evaluation dimension is defined in a prompt template under one of four paradigms: (i) Binary (deterministic pass/fail; e.g., ``Did the patient mention self-harm?''); (ii) Scalar (Likert ratings; e.g., ``Rate the consistency from 1 to 5.''); (iii) Categorical (labels; e.g., ``How realistic are the patient's responses? \texttt{good}/\texttt{neutral}/\texttt{bad}''); and (iv) Extraction (evidence spans; e.g., ``Highlight statements that contradict the given profile.'').
This abstraction facilitates instruction customization while keeping the interface uniform across tasks and methods.

\subsection{Usage}
PatientHub's workflow is accessible through a single command-line interface. 
Users can \texttt{generate} profiles from seed records, \texttt{simulate} an interaction between agents, \texttt{evaluate} sessions or profiles with a judge, and \texttt{create} a new agent implementation that becomes immediately available to the framework. 
Moreover, as patient simulators typically define profiles in incompatible formats, reusing a persona across methods otherwise requires manual reconstruction. 
Our framework includes an \texttt{adapt} command that maps a persona to another method's schema.
By automating this conversion, PatientHub enables a single persona to drive heterogeneous simulators and supports controlled cross-method comparison.
Hydra can override all settings, and an optional web interface supports interactive, human-in-the-loop sessions.

\subsection{Technical Details}
PatientHub is implemented in Python with an emphasis on reproducibility and modularity.
We use Hydra to manage configurations across datasets, agents, and experimental settings, Burr to represent and execute events as graphs, and LiteLLM as a unified interface for LLM calls.
Structured outputs are validated with Pydantic schemas, and prompts are authored as Jinja-rendered YAML templates over JSON character records.
New components are added by inheriting a base class (e.g., \texttt{BaseClient}) and registering it in the framework, at which point it becomes immediately available to the CLI.

\section{Evaluation}
We demonstrate PatientHub's utility through a single end-to-end workflow over a shared set of personas: we i) create client profiles through generation and adaptation, validating them with the profile judge; (ii) simulate and compare several methods under a common protocol, scored by the conversation judge; and (iii) extend the framework with a new method. 
Each stage demonstrates a core capability; we aim to show that PatientHub standardizes and executes the full pipeline, not to benchmark simulators or identify a single best method.

\subsection{Experimental Setup}
\paragraph{Therapists.} 
We adopt Cognitive Behavioral Therapy (CBT), one of the most extensively studied and widely practiced forms of psychotherapy \cite{david2018cognitive, hofmann2012efficacy}, to instantiate the therapist agent (CBT therapist; Figure \ref{fig:cbt-prompt}).
Moreover, to probe client simulator behavior under adverse conditions, we define a \emph{bad} therapist (dismissive, invalidating, and lacking empathy) that serves as a known-contrast stress condition rather than a realistic counselor (Figure \ref{fig:bad-prompt}).

\paragraph{Clients.}
We select four simulators spanning our supported paradigms and run them under a CBT protocol: single-agent prompting (PATIENT-$\Psi$, CARS), multi-agent prompting (MindVoyager), and fine-tuning (Eeyore).
This selection also covers static and dynamically evolving clients (e.g., CARS's turn-level resistance). 
To ensure all methods are driven by the same underlying cognitive models, we use the human-validated cognitive conceptualization diagrams (CCDs) from PATIENT-$\Psi$ \cite{wang-etal-2024-patient} as the persona seed (50 in total).
Each method is instantiated from the same CCD: PATIENT-$\Psi$ uses them directly, MindVoyager reuses the same cases grouped by individual, CARS expands each CCD through its generator, and Eeyore is instantiated via the \texttt{adapt} command.

\paragraph{Hyperparameters.}
All prompted agents are instantiated with GPT-4o \cite{hurst2024gpt} as the backbone LLM; for Eeyore, we use the fine-tuned Llama-3.1-8B model provided by the original authors\footnote{\url{https://huggingface.co/liusiyang/eeyore_sft_epoch2_dpo_round2_epoch1_llama3.1_8B}}.
We used Temperature = $0.7$ for all agents.

\subsection{Profile Creation}
Users can create profiles in two ways and leverage the profile judge to validate both.
In our experiments, we employ the \emph{extraction} paradigm, which flags concrete issues (missing context or inconsistencies) and suggests revisions rather than producing a single opaque rating (Figures \ref{fig:profile_judge_gen} \& \ref{fig:profile_judge_adapt}).

First, several methods provide generators that can be invoked via \texttt{generate}; we demonstrate this with CARS, whose generator expands a CCD into a full CARS profile.
Applying the judge to the generated profiles consistently surfaced completeness gaps (e.g., unpopulated schema fields and under-specified history) and coherence contradictions, such as a client whose coping strategy of seeking social support conflicts with her described isolation and belief that she must cope alone (Table~\ref{tab:profile_case_study}), localizing each issue for targeted revision.

Second, the \texttt{adapt} command converts a persona across method schemas, enabling a single persona to drive otherwise-incompatible simulators.
As Eeyore is the only selected method that is not CCD-based, we adapt each CCD to its schema and infer target-specific fields (e.g., demographics and severity ratings) that the source leaves unspecified.
The judge results indicated that adaptation introduced no contradictions with the source facts in most cases (43/50), with the few flagged discrepancies being minor re-framings or differences in inferred severity rather than contradictions (e.g., assigning a ``mild to moderate'' depression severity where the source describes an intensifying low mood without naming depression).
We manually revised each profile to resolve the issues the judge flagged before using it to simulate conversations.

\begin{table}[t]
\caption{Profile judge feedback (paraphrased) on a generated profile. The judge localizes coherence and completeness issues, surfacing a contradiction between the client's stated coping strategy and her described isolation and belief that she must cope alone.}
\label{tab:profile_case_study}
\centering
\small
\begin{tabular}{p{\columnwidth}}
  \toprule
  \textbf{Snippet from a Generated CARS Profile (Grace)}\\
  \midrule
  \begin{minipage}[t]{\linewidth}
    \ttfamily
    \textbf{Coping Strategies:} Seeks social support and spends time with friends; keeps busy and avoids difficult situations. \\
    \textbf{Background:} Following her divorce, Grace isolates herself and has difficulty trusting people. \\
    \textbf{Intermediate Beliefs:} ``I must handle everything alone and never ask for help.'' / ``If people know my situation, they will not want to support me.'' \\
    \textbf{Session CCD:} \texttt{triggers}: financial stress, social gatherings. \\
    \textbf{Preferences:} Values a counselor who validates her feelings and acknowledges her struggles. \\
    \vspace{-0.6em}
  \end{minipage}
  \\
  \midrule
  \textbf{Profile Judge Feedback}\\
  \midrule
  \begin{minipage}[t]{\linewidth}
    \textbf{Coherence:} Coping strategies list ``seeking social support and spending time with friends,'' yet the background states she isolates herself and struggles to trust people after her divorce. This also conflicts with her belief that she ``must handle everything alone,'' and with her stated preference for a counselor who validates her, despite a described tendency to withdraw. \\[2pt]
    \textbf{Completeness:} The coping strategies ``keeping busy and avoidance'' are vague, giving no specifics of what avoidance entails. The two intermediate beliefs are near duplicates and lack nuance, and the session-CCD triggers are listed without explaining their significance. \\
    \vspace{-0.6em}
  \end{minipage}\\
  \bottomrule
\end{tabular}
\end{table}

\begin{table*}[t]
\caption{Cross-method comparison under the CBT and bad-therapist conditions, scored by an independent judge (Sonnet-4.6) over 50 personas per condition. Ratings are on a 1--5 Likert scale; response length is in tokens and API cost in USD, both averaged per session (Eeyore is self-hosted and incurs no API cost).}
\centering
\begin{tabular}{llccccc}
\toprule
\textbf{Therapist} & \textbf{Aspect} & \textbf{PATIENT-$\Psi$} & \textbf{CARS} & \textbf{MindVoyager} & \textbf{Eeyore} & \textbf{$\Psi$-Doh} \\
\midrule
\multirow{9}{*}{\textbf{CBT}}
& Factual Consistency     & 3.94 & 3.94 & 3.72 & 3.80 & 3.94 \\
& Cross-turn Coherence    & 4.02 & 3.94 & 3.94 & 3.86 & 3.86 \\
& Psychological Alignment & 3.94 & 3.90 & 3.62 & 3.62 & 3.94 \\
& Naturalness             & 3.04 & 3.06 & 3.92 & 2.96 & 3.66 \\
& Emotional Depth         & 3.28 & 3.14 & 3.28 & 2.64 & 3.20 \\
& Appropriate Resistance  & 2.44 & 3.70 & 2.44 & 2.60 & 3.22 \\
& Feedback Quality        & 3.14 & 3.22 & 3.14 & 2.80 & 3.46 \\
\cmidrule(lr){2-7}
& Response Length         & 86.0 & 26.0 & 96.5 & 14.8 & 53.0 \\
& API Cost                & \$0.05 & \$0.09 & \$0.11 & N/A&\$0.15 \\
\midrule
\multirow{9}{*}{\textbf{Bad}}
& Factual Consistency     & 3.86 & 3.34 & 2.72 & 3.62 & 3.86 \\
& Cross-turn Coherence    & 3.94 & 3.78 & 3.94 & 3.26 & 3.94 \\
& Psychological Alignment & 4.02 & 3.42 & 3.52 & 3.42 & 3.94 \\
& Naturalness             & 3.94 & 3.22 & 3.68 & 3.14 & 2.96 \\
& Emotional Depth         & 3.22 & 2.96 & 3.28 & 2.78 & 2.96 \\
& Appropriate Resistance  & 2.88 & 3.74 & 2.70 & 3.04 & 3.28 \\
& Feedback Quality        & 2.70 & 3.74 & 3.30 & 3.12 & 3.30 \\
\cmidrule(lr){2-7}
& Response Length         & 80.5 & 21.0 & 74.0 & 10.6 & 30.5 \\
& API Cost                & \$0.08 & \$0.05 & \$0.09 & N/A&\$0.15 \\
\bottomrule
\end{tabular}
\label{tab:eval_res}
\end{table*}

\subsection{Cross-method Comparison}
We pair each simulator with both therapists using the \emph{Therapy Session} event (15 turns, with a moderator prompting closure at turn 13), running one session per persona (50 per therapist).
To reduce self-preference bias, we use an independent LLM (Sonnet-4.6) as the conversation judge (Figure \ref{fig:conv_judge}) rather than the GPT-4o backbone used by the simulators. 
As all methods are instantiated from the same CCDs, the framework runs heterogeneous simulators under a single protocol and surfaces their differences directly, a comparison that would be implausible with incompatible profile formats. 

Prior work mainly assesses client performance via three dimensions: consistency \cite{wang2024towards, abdulhai2025consistently}, realism \cite{lee2025psyche, kim-etal-2025-share-story}, and pedagogical utility \cite{wang-etal-2024-patient, louie2025llmsimulated}.
To assess consistency, we include \emph{factual consistency} (stated facts match the profile), \emph{cross-turn coherence} (no contradictions across turns), and \emph{psychological alignment} (expressed beliefs and reactions align with the profile).
Moreover, we include \emph{naturalness} (human-like expression), \emph{emotional depth}, and \emph{appropriate resistance} (the client is not overly forthcoming) to assess realism. 
Lastly, to assess pedagogical utility, we evaluate \emph{feedback quality} by whether client responses provide trainees with useful signals.
We also report the average response length and API cost per session for each method. 

On the judged dimensions (Table~\ref{tab:eval_res}), all methods score highly on consistency under the cooperative CBT therapist and thus appear broadly similar. 
The adversarial \emph{bad} therapist serves as a stress test that separates otherwise similar methods. 
Notably, CARS, the only method that can end a session itself, disengages early, averaging just 5.1 of the 15 allotted turns, effectively walking out of an unproductive session, which aligns with it having the highest \emph{appropriate resistance} scores under both therapists. 
That appropriate resistance rises under the bad therapist for every method is a sanity check that the judge tracks interaction quality rather than surface form. 
The realism dimensions, by contrast, are illustrative: prior work finds LLM judges can diverge from or even invert expert ratings on exactly these fidelity dimensions \cite{wang-etal-2024-patient, hoang2026psibench}, so we make no claim they capture clinical realism.
Beyond the scores, the framework exposes efficiency differences that per-method reports leave implicit. 
Per-session cost tracks architectural overhead (e.g., extra agent calls and intermediate reasoning tokens) rather than response length: the multi-agent and reasoning-augmented methods incur the highest cost despite not producing the longest replies, while Eeyore runs locally at no API cost.
For users selecting a simulator at scale, these trade-offs are decisive.

\subsection{Extensibility}
We leverage the \texttt{create} command to prototype $\Psi$-Doh, a PATIENT-$\Psi$ variant that reuses its profile schema and adds turn-level reasoning based on a therapist-trust estimate \cite{srivastava2025trust} that modulates the level of cooperation and disclosure, and performs turn-level self-evaluation and response rewriting, following \cite{louie-etal-2024-roleplay}.
$\Psi$-Doh requires only 72 lines of code on top of PATIENT-$\Psi$: 45 for Pydantic schemas and 27 for the trust-estimation and rewriting logic.

As shown in Table~\ref{tab:eval_res}, $\Psi$-Doh exhibits measurably different behavior from base PATIENT-$\Psi$, primarily higher appropriate resistance and feedback quality across both therapists, at shorter response lengths but with higher per-session cost, reflecting its added reasoning and rewriting steps. 
As these overlap with the illustrative fidelity dimensions above, we read them as behavioral differences rather than improvements; we show that a new method integrates seamlessly with PatientHub and is immediately comparable to existing work.

\section{Conclusion}
We presented PatientHub, a unified and modular framework that standardizes the creation, simulation, and evaluation of simulated patients.
Through a single command-line interface, PatientHub consolidates various simulation methods under shared abstractions for agents, events, generators, and rubric-driven evaluation.
We evaluated PatientHub's capabilities end-to-end, showing that it lowers the barrier to building, comparing, and extending patient simulators: creating profiles through generation and adaptation, comparing simulators under a common protocol, and prototyping a new method with minimal code.
PatientHub is open-source, and we will continue to expand its coverage of simulation methods and clinical domains.

\section*{Limitations}
Our evaluation demonstrates PatientHub's capabilities rather than providing a comprehensive benchmark. 
It covers a limited set of personas (50 CCDs) within a single therapeutic domain (CBT counseling) and is not intended to establish which simulator is best or to characterize method performance at scale.
In addition, we demonstrate single-session, multi-turn counseling; PatientHub does not yet target multi-party care teams, long-horizon follow-up, crisis intervention, or tool-augmented clinicians.

To reduce self-preference bias, we used Sonnet-4.6 as the conversation judge, which is distinct from the simulators' backbone (GPT-4o).
However, rubric-driven LLM-as-a-judge evaluation carries known concerns: judgments can be sensitive to rubric phrasing and may correlate with surface features (e.g., response length). 
Prior work reports that LLM-judge fidelity scores can align weakly with, or even invert, expert ratings \cite{wang-etal-2024-patient, hoang2026psibench, sabour2026patientact}.
Hence, we rely on extraction-based checks for our conclusions and treat the Likert realism dimensions as illustrative; validating them against expert clinicians remains important future work.

Finally, adaptation carries a further caveat. 
As Eeyore lacks a profile generator, we instantiate its profiles by adapting personas across schemas, so our cross-method comparison reflects a shared persona seed rather than identical inputs.
We also report contradictions rather than a full preservation score, as we found LLM-judge fact-matching too unreliable for the latter (e.g., it conflated schema-driven omissions and paraphrases with genuine contradictions); therefore, we verified adapted profiles manually, and automating faithful cross-schema adaptation remains an open problem.

\section*{Ethics Statement}
PatientHub is a research framework that operates without requiring real patient records. 
If users supply their own data (e.g., conversation logs), they are responsible for obtaining appropriate consent, de-identifying the data, and complying with applicable laws and institutional review board requirements.
Simulated interactions may produce unsafe content (e.g., harmful advice, inappropriate escalation, or privacy leakage), and automatically generated personas may encode or amplify societal biases.
Moreover, automatically generated or adapted profiles may include clinically sensitive attributes (e.g., symptom severity or risk indicators such as suicidal ideation) inferred by an LLM and not clinically validated; these should never be treated as real clinical assessments. 
We therefore recommend using PatientHub in controlled settings, supervising profile and dialogue generation, and not deploying LLM-based supporters to real users without expert evaluation.
If PatientHub is used in studies involving human participants (e.g., trainees or clinicians), evaluations should follow established human-subjects protocols and prioritize participant well-being.

\section*{Acknowledgments}
This work was supported by the National Science Foundation for Distinguished Young Scholars (\# 62125604), the National Key Research and Development Program of China (\#2024YFC3606800), the NSFC projects (\#62441614), and the Beijing Natural Science Foundation (\#L252009).

\bibliography{citations}

\appendix
\onecolumn
\section{Supported Client Simulators}

\begin{table*}[htbp]
\caption{List of patient simulators implemented in PatientHub. CBT: Cognitive Behavioral Therapy; MI: Motivational Interviewing; GP: General Psychotherapy; and CM: Clinical Medicine.}
\centering
\begin{tabularx}{\linewidth}{lllll}
\hline
\textbf{Method} & \textbf{Domain} & \textbf{Task} & \textbf{Implementation} & \textbf{Interaction} \\
\hline
\methodlink{PATIENT-$\Psi$}{wang-etal-2024-patient} & CBT & Counselor Training & Single-agent prompting & Multi-turn \\
\methodlink{MindVoyager}{kim-etal-2025-share-story} & CBT & Counselor Training & Multi-agent prompting & Multi-turn \\
\methodlink{CARS}{qin2026client} & CBT & Counselor Training & Single-agent prompting & Multi-turn \\
\methodlink{Eeyore}{liu-etal-2025-eeyore} & GP & Counselor Training & SFT + DPO Fine-tuning & Multi-turn \\
\methodlink{Roleplay-doh}{louie-etal-2024-roleplay} & GP & Counselor Training & Multi-agent prompting & Multi-turn \\
\methodlink{ClientCAST}{wang2024towards} & GP & Therapy Evaluation & Single-agent prompting & Multi-turn \\
\methodlink{AnnaAgent}{wang-etal-2025-annaagent} & GP & Realistic Simulation & Multi-agent prompting & Multi-session \\
\methodlink{Deprofile}{li2026synthetic} & GP & Realistic Simulation & Single-agent prompting & Multi-turn \\
\methodlink{SimPatient}{steenstra2025scaffolding} & MI & Counselor Training & Multi-agent prompting & Multi-session \\
\methodlink{Consistent MI}{yang-etal-2025-consistent} & MI & Realistic Simulation & Multi-agent prompting & Multi-turn \\
\methodlink{PSYCHE}{lee2025psyche} & GP & Psychiatric Diagnosis & Single-agent prompting & Multi-turn \\
\methodlink{TalkDep}{wang2025talkdep} & GP & Depression Diagnosis & Single-agent prompting & Multi-turn \\
\methodlink{SAPS}{liao2024automatic} & CM & Depression Diagnosis & Multi-agent prompting & Multi-turn \\
\methodlink{Adaptive-VP}{lee-etal-2025-adaptive} & CM & Communication Training & Multi-agent prompting & Multi-turn \\
\methodlink{PatientZero}{lai2026patientzero} & CM & Realistic Simulation & Single-agent prompting & Multi-turn \\
\methodlink{PatientAct}{sabour2026patientact} & CM & Realistic Simulation & Multi-agent prompting & Multi-turn \\
\hline
\end{tabularx}

\label{tab:supported_clients}
\end{table*}

\section{System Prompts}
\begin{figure*}[htbp]
\begin{promptbox}
You are an AI assistant role-playing as a skilled Cognitive Behavioral Therapy therapist. 
Your primary goal is to guide the user through a supportive conversation grounded in CBT principles.

\# Rules of Engagement

- Your task is to help the user explore their own thoughts and feelings to reach their own insights.

- You will never provide a medical diagnosis. Your focus is on exploring thought patterns, emotions, and behaviors in the present moment.

- Briefly summarize what you hear and acknowledge their emotion (e.g., "It sounds like you're feeling [emotion] because [situation]. That must be really difficult.").

- Help the user separate the situation, their thoughts about it, and the resulting emotions. Use questions like, "When [situation] happens, what goes through your mind?"

- Use open-ended questions to explore and gently challenge unhelpful thoughts. e.g., "What is the evidence for and against that thought?" and "Is there an alternative way to see this situation?"

- Based on the exploration, collaboratively help the user develop a more balanced or helpful perspective. Ask, "Based on our discussion, could we try to phrase a new, more balanced thought?"

- (if appropriate) Connect new thoughts to small, actionable steps. Ask, "What's a small thing you could do that aligns with this new perspective?"

\# Requirements

- Keep responses concise (maximum 3 sentences) and focused on one step at a time.

- Use empathy-focused language ("I understand," "That makes sense").

- Always end a response with an open-ended question to continue the collaborative exploration.

- Avoid giving direct advice or definitive answers. Your role is to question and guide.

- Begin the session by asking: "Hello, what's on your mind today?"

\end{promptbox}
\caption{System Prompt for CBT Therapist.}
\label{fig:cbt-prompt}
\end{figure*}

\begin{figure*}[htbp]
\begin{promptbox}
In this task, you will be role-playing as a therapist who is unprofessional, dismissive, and lacks empathy towards your clients. 
Your responses should reflect a lack of understanding and compassion, often invalidating the client's feelings and experiences. 

You may also provide inappropriate advice or make light of serious issues. Remember to maintain a tone that is cold, indifferent, and sometimes sarcastic. Your goal is to create a challenging and uncomfortable experience for the client, highlighting what not to do in a therapeutic setting.

With that in mind, please respond to the client's statements in a way that is unhelpful and unsupportive.
Keep your responses brief and to the point, avoiding any genuine engagement with the client's concerns.
Your responses will be used to evaluate the quality of simulated patients, so please ensure they are consistently unprofessional and dismissive throughout the interaction.
\end{promptbox}
\caption{System Prompt for Bad Therapist.}
\label{fig:bad-prompt}
\end{figure*}

\begin{figure*}[htbp]
\begin{promptbox}
You are an expert evaluating the quality of a generated client profile used to simulate a standardized patient (SP) in CBT therapy training. These profiles are used to program simulated patients for training therapists.

\# Character Profile: [profile]

Your task is to identify specific issues, errors, or weaknesses in the profile based on the provided criteria, extracting the exact problematic spans rather than assigning an overall score.
\end{promptbox}
\caption{System Prompt for Profile Judge (for generated profiles).}
\label{fig:profile_judge_gen}
\end{figure*}

\begin{figure*}[htbp]
\begin{promptbox}
You are an expert at checking whether a client profile adapted from a source schema to a target schema remains faithful to the source. These profiles simulate standardized patients for CBT therapy training.

\# Source Profile:[source]

\# Adapted Profile: [profile]

Extract only statements in the adapted profile that conflict with a stated fact in the source. 

A contradiction requires the source to explicitly assert the opposite; restatements, paraphrases, and elaborations are not contradictions, nor is the absence of a statement a contradiction. 

The target schema may contain fields the source does not address; inferring plausible values for these fields (e.g., demographics or severity ratings) is expected and must not be flagged. 

For each contradiction, you MUST quote the exact source sentence being contradicted.

\end{promptbox}
\caption{System Prompt for Profile Judge (for adapted profiles).}
\label{fig:profile_judge_adapt}
\end{figure*}

\begin{figure*}[htbp]
\begin{promptbox}
  You are an expert at evaluating whether a simulated patient (SP) playing the role of a client in a CBT therapy session faithfully adheres to the prescribed dimensions and their aspects.

  \# Profile: [profile]

  \# Conversation History: [conversation history]

  Your task is to rate the SP's performance on each aspect using the provided scale.
  You should focus on the SP's performance across the entire session.
  The target of this evaluation is the client. You should only assess responses made by the client.

  IMPORTANT: Use the full 1-5 scale with the following behavioral anchors:

  1 – Poor: The SP clearly fails on this aspect. Errors are obvious, frequent, or significantly break believability/consistency.
  
  2 – Below average: Noticeable weaknesses that a trained observer would flag. The SP partially meets expectations but has clear gaps.
  
  3 – Adequate: Meets the basic requirement but in a mechanical or unremarkable way. No notable failures, but also no nuance or depth.
  
  4 – Good: Clearly above average with only minor, easily-overlooked shortcomings. The SP demonstrates intentionality and skill.
  
  5 – Exceptional: Near-flawless execution that a domain expert would consider exemplary.

  Default to 3 when in doubt. A score of 4 requires a positive reason; a score of 5 requires an exceptional reason. Most SPs will score 2–3 on at least some aspects.
\end{promptbox}
\caption{System Prompt for the Conversation Judge.}
\label{fig:conv_judge}
\end{figure*}

\end{document}